# AdaEnsemble Learning Approach for Metro Passenger Flow Forecasting

Shaolong Sun[a], Dongchuan Yang[a], Ju-e Guo[a], Shouyang Wang[b,c,d,*]

[a]School of Management, Xi'an Jiaotong University, Xi'an, 710049, China

[b]Academy of Mathematics and Systems Science, Chinese Academy of Sciences, Beijing 100190, China

[c]School of Economics and Management, University of Chinese Academy of Sciences, Beijing 100190, China

[d]Center for Forecasting Science, Chinese Academy of Sciences, Beijing 100190, China

*Corresponding author. Academy of Mathematics and Systems Science, Chinese Academy of Sciences, Beijing 100190, China. Tel.: +86 10 82541772; fax: +86 10 82541972.

E-mail address: sywang@amss.ac.cn (S. Y. Wang).

**Abstract:** Accurate and timely metro passenger flow forecasting is critical for the successful deployment of intelligent transportation systems. However, it is quite challenging to propose an efficient and robust forecasting approach due to the inherent randomness and variations of metro passenger flows. In this study, we present a novel adaptive ensemble (AdaEnsemble) learning approach to accurately forecast the volume of metro passenger flows that combines the complementary advantages of variational mode decomposition (VMD), seasonal autoregressive integrated moving averaging (SARIMA), a multilayer perceptron (MLP) network and a long short-term memory (LSTM) network. The AdaEnsemble learning approach consists of three important stages. The first stage applies VMD to decompose the metro passenger flow data into periodic components, deterministic components and volatility components. Then, we employ the SARIMA model to forecast the periodic component, the LSTM network to learn and forecast the deterministic component and the MLP network to forecast the volatility component. In the last stage, these diverse forecasted components are reconstructed by another MLP network. The empirical results show that our proposed AdaEnsemble learning approach not only has the best forecasting performance compared with the state-of-the-art models but also appears to be the most promising and robust based on the historical passenger flow data in the Shenzhen subway system and several standard evaluation measures.

## 1 Introduction

Metro transportation systems have played a vital role in urban traffic configurations. They not only provide a means of reducing ground traffic congestion and delays but also offer the advantages of high safety, reliability and efficiency, and they have become increasingly popular. There were approximately 5.1 million metro trips every day in Shenzhen in 2018, accounting for 48% of the total public passenger flow. Passenger flow forecasting is a critical component in an urban metro system because it is critically important to develop a reasonable operating plan to match transport capacity and passenger demand, fine-tune passenger travel behaviors, improve transport services and reduce the level of congestion. In the field of transportation, research on metro passenger flow forecasting has attracted increasing attention and can be categorized as studying short-term, medium-term and long-term issues; the short-term issue is foremost in extant research.

Metro passenger flow tends to have daily, weekly and seasonal periodic patterns, and the pedestrian movement patterns of passengers on weekdays and weekends are completely different (Diao et al., 2019; Ke et al., 2017; Wei and Chen, 2012). The majority of passengers regularly take metros as commuter vehicles on weekdays, while on weekends, metros are randomly used (Sun et al., 2015). Furthermore, the pedestrian movement patterns of passengers are sensitive to special events, extreme weather conditions, accidents, etc., and they may slightly adjust their travel time, transferring stations and mode choice to avoid rush hours (Zhang et al., 2015). Hence, short-term metro passenger flow forecasting is a hard issue, and there is still much to do to improve the accuracy of short-term traffic forecasting, which is a critical element in traffic systems (Chan et al., 2012; VanArem et al., 1997; Wang et al., 2014).

The change in metro passenger flow is a real-time, nonlinear and nonstationary random process. With the shortening of the statistical period, the metro passenger flow becomes more uncertain and random. The time series of metro passenger flow obviously has characteristics of temporal periodicity, high fluctuation and nonlinearity. Therefore, it is difficult to accurately predict metro passenger flow using linear or nonlinear models alone (Bai et al., 2017; Zhang and Haghani, 2015). Because of the temporal periodicity, high volatility and nonlinearity of metro passenger flow, decomposing the metro passenger flow and using a hybrid model for prediction is an effective solution.

The purpose of this paper is to resolve this gap in the literature with variational mode decomposition (VMD) to obtain periodic components, deterministic components and volatility components. We use the SARIMA model to predict the periodic component, use the LSTM network to learn and predict the deterministic component, and use the MLP network to predict the volatility component. In the final stage, various prediction components are reconstructed through another MLP network. To verify the superiority of our proposed AdaEnsemble learning method, we established five predictive models (i.e., seasonal autoregressive integrated moving averaging (SARIMA) model, multilayer perceptron (MLP) neural network, long short-term memory (LSTM) network, and two decomposition ensemble learning approaches including VMD-MLP and VMD-LSTM) and used them as benchmarks to make multistep prediction comparisons of three Shenzhen subway stations.

The rest of this study is organized as follows: a comprehensive literature review is provided in **Section 2**. Then, the related methodology is introduced in **Section 3**. The empirical results and performance of our proposed approach are discussed in **Section 4**. Finally, concludings and suggestions for future work are offered in **Section 5**.

## 2 Literature review

Over the past few decades, short-term traffic forecasting has attracted widespread attention from worldwide researchers. Generally, traffic forecasting models can be divided into two major categories: parametric models and nonparametric models. In addition, hybrid models and decomposition techniques are also widely used in short-term traffic forecasting. Each family of the above models is described in detail below.

First, in a variety of parametric models, many prototypes of different models have been proposed for traffic flow forecasting, such as moving average models, exponential smoothing models, gray forecasting models, autoregressive integrated moving average (ARIMA) models (Hamzaçebi, 2008; Tsui et al., 2014), and state space models (Stathopoulos and Karlaftis, 2003). ARIMA is a linear combination of time lagged variables, which has become one of the widely used parametric forecasting models since the 1970s because it performs well and effectively in modeling linear and stationary time series. However, it may not capture the structure of nonlinear relationships due to the assumption of linear relationships among time lagged variables (Zhang, 2003). Other models also have their shortcomings, and gray forecasting models may cause large deviations of forecast results due to the sparse and volatile samples.

Second, in the family of nonparametric models, numerous approaches have been applied to forecast traffic flow, including nonparametric regression methods such as Gaussian maximum likelihood (Tang et al., 2003), artificial neural networks (Chen et al., 2012; Tsai et al., 2009), support vector regression (Chen et al., 2012; Sun et al., 2015; Wu et al., 2004; Yao et al., 2017), and other models (Dumas and Soumis, 2008; Sun, 2016). Among these nonparametric models, artificial neural networks have gained much research interest for passenger flow forecasting because of their adaptability, nonlinearity, arbitrary functions and mapping capabilities (Vlahogianni et al., 2004). Artificial neural network applications extend from the simple multilayer perceptron to complex structures such as wavelet-based neural networks (Boto-Giralda et al., 2010), Kalman filtering-based multilayer perceptron (Lippi et al., 2013), Jordan's sequential neural networks (Yasdi, 1999), finite impulse response networks (Yun et al., 1998), time-delayed recurrent neural networks, dynamic neural networks (Ishak and Alecsandru, 2004), Elman neural networks (Chen and Grant-Muller, 2001), and spectral basis neural networks (Park et al., 1999). However, neural networks also have some intrinsic drawbacks, such as the local minima issue, the selection of the number of hidden units and the danger of overfitting. Additionally, to obtain a good generalization performance, larger in-samples are needed. Cortes and Vapnik (1995) proposed another widely used nonparametric model named support vector machine (SVM), which is based on the principle of structural risk minimization (minimizing an upper bound on the generalization error). SVM has the potential to overcome the shortcomings of neural networks and can be skilled in the problems of nonlinearity, small samples, high dimensionality, local minima and overfitting.

Third, hybrid models have been demonstrated to provide better performance than single models in traffic flow forecasting, including a hybrid model that combines both ARIMA and a multilayer artificial neural network, genetic algorithms and a gray model combined with a SVM (Jiang et al., 2014) , nonlinear vector auto-regression neural network combined with mean impact value (Sun et al., 2019), variational mode decomposition (VMD) and ARMA combined with kernel extreme learning machine (KELM) (Jin et al., 2020). Recently, Ni et al (2017). proposed an ensemble learning framework to appropriately combine estimation results from multilayer macroscopic traffic flow models. This framework assumed that any existing models were imperfect and had their own weaknesses/strengths, but the ensemble learning framework enabled the combination of every individual estimation model to improve traffic state estimation accuracy.

Fourth, to better capture traffic characteristics, numerous approaches have been applied to decompose traffic flow into different components, including wavelet transform, empirical mode decomposition and seasonal decomposition. Empirical mode decomposition is a powerful multiresolution signal decomposition technique. It is an empirical, direct and adaptive data processing method that is appropriate for dealing with nonlinear and nonstationary data. Wei and Chen predicted metro passenger flows with a hybrid of EMD and neural networks that generated higher forecasting accuracy and stability than the seasonal ARIMA (Wei and Chen, 2012). Wavelet decomposition is an effective way of analyzing the passenger flow data in both time and frequency domains. Diao et al. (2019) decomposed a traffic volume series into several components by discrete wavelet transform and predicted different components with a Gaussian process model and a tracking model. Seasonal decomposition is an effective method to decompose time series into trend components, seasonal components and irregular components based on seasonal decomposition and least squares support vector regression (LSSVR). Xie et al. (2014) proposed two hybrid approaches to conduct a short-term forecasting of air passengers. Variational mode decomposition (VMD) is a novel nonrecursive and adaptive signal decomposition algorithm. Li et al. (2020) decomposed an air cargo time series by an enhanced decomposition formwork, which consists of sample entropy (SE), empirical mode decomposition (EMD) and variational mode decomposition (VMD). Niu et al. (2018) decomposed container throughput time series into low-frequency components and high-frequency components by Variational mode decomposition (VMD).

## 3 Related methodology

In this section, before presenting our proposed AdaEnsemble learning approach, we first introduce some methods that will be applied in our approach.

### 3.1 Variational mode decomposition

Variational mode decomposition (VMD), originally proposed by Dragomiretskiy and Zosso (2014), is a novel nonrecursive and adaptive signal decomposition algorithm that can accommodate much more sampling and noise than popular decomposition techniques such as wavelet transform (WT) and empirical mode decomposition (EMD). The main goal of VMD is to decompose an original signal into a discrete set of band-limited modes $u_k$, where each mode $u_k$ is considered to be mostly compact around a center pulsation $\omega_k$, which is determined during the decomposition. The bandwidth of each mode $u_k$ is estimated though the following scheme:

**Step 1**: Apply the Hilbert transform to calculate the associated analytical signal for each mode $u_k$ to obtain a unilateral frequency spectrum.

**Step 2**: Shift the frequency spectrum of each mode to the baseband by means of mixing with an exponential tuned to the respective estimated center frequency.

**Step 3**: Estimate the bandwidth of each mode $u_k$ through the Gaussian smoothness of the demodulated signal.

For instance, the time series $f$ is decomposed into a set of modes $u_k$ around a center pulsation $\omega_k$ according to the following constrained variational problem:

$$\min_{\mu_k,\omega_k} \sum_k \left\| \partial_t \left[ \left( \delta(t) + \frac{j}{\pi t} \right) * \mu_k(t) \right] e^{-j\omega_k t} \right\|_2^2 \tag{1}$$

Subject to

$$\sum_k u_k = f \tag{2}$$

where $\delta$ is the Dirac distribution, $k$ is the number of modes, and $*$ is the convolution operator. $\{u_k\}$ and $\{\omega_k\}$ represent the set of modes $\{u_1, u_2, ..., u_k\}$ and the set of center pulsations $\{\omega_1, \omega_2, ..., \omega_k\}$, respectively.

In the VMD framework, the original time series $f$ is decomposed into a set of modes $u_k$ around a center pulsation $\omega_k$, and each has a bandwidth in the Fourier domain (see equation (1)). The solution to the above constraint variational problem can be headed with an unconstrained optimization problem according to a quadratic penalty term and Lagrange multipliers $\lambda$, which is given as follows:

$$L(\mu_k,\omega_k,\lambda) = \alpha \sum_k \left\| \partial_t \left[ \left( \delta(t) + \frac{j}{\pi t} \right) * \mu_k(t) \right] e^{-j\omega_k t} \right\|_2^2 + \left\| f(t) - \sum_k \mu_k(t) \right\|_2^2 + \left\langle \lambda(t), f(t) - \sum_k \mu_k(t) \right\rangle \tag{3}$$

where $\alpha$ represents a balance parameter of the data fidelity constraint, $\lambda$ represents the Lagrange multipliers, and $\left\| f(t) - \sum_k \mu_k(t) \right\|_2^2$ denotes a quadratic penalty term for the accelerating rate of convergence.

Furthermore, the solution to Eq. (2) can be solved by the alternative direction method of multipliers (ADMM) by means of finding the saddle point of the augmented Lagrangian function $L$ in a sequence of iterative suboptimizations. Consequently, the solutions for $\mu_k$, $\omega_k$ and $\lambda$ can be obtained as follows:

$$\hat{\mu}_k^{n+1}(\omega) = \frac{\hat{f}(\omega) - \sum_{i \neq k} \hat{\mu}_i(\omega) + \frac{\hat{\lambda}(\omega)}{2}}{1 + 2\alpha(\omega - \omega_k)^2} \tag{4}$$

$$\omega_k^{n+1} = \frac{\int_0^\infty \omega |\hat{\mu}_k(\omega)|^2 d\omega}{\int_0^\infty |\hat{\mu}_k(\omega)|^2 d\omega} \tag{5}$$

$$\hat{\lambda}^{n+1}(\omega) = \hat{\lambda}^n(\omega) + \tau\left(\hat{f}(\omega) - \sum_k \hat{\mu}_k^{n+1}(\omega)\right) \tag{6}$$

where $\hat{f}(\omega)$, $\hat{\mu}_i(\omega)$, $\hat{\lambda}(\omega)$, $\hat{\lambda}^n(\omega)$ and $\hat{\mu}_k^{n+1}(\omega)$ represent the Fourier transforms of $f(\omega)$, $\mu_i(\omega)$, $\lambda(\omega)$, $\lambda^n(\omega)$ and $\mu_k^{n+1}(\omega)$, respectively, and $n$ is the number of iterations.

Before the VMD method, the number of modes $k$ should be determined. The mode $\mu$ with high order $k$ represents low-frequency components. There is no theory regarding optimal selection of the parameter $k$. In this study, its value is set to 3. For further details on the VMD algorithm, please refer to Dragomiretskiy and Zosso (2014).

### 3.2 Seasonal autoregressive integrated moving average

A time series $\{X_t\}$ is a seasonal $ARIMA(p,d,q)(P,D,Q)_S$ process with period $S$ if $d$ and $D$ are nonnegative integers and if the differenced series $Y_t = (1-B)^d (1-B^S)^D X_t$ is a stationary autoregressive moving average (ARMA) process. It can be expressed by:

$$\phi(B)\Phi(B^S)Y_t = \theta(B)\Theta(B^S)\varepsilon_t \tag{7}$$

where $B$ is the backshift operator defined by $B^a X_t = X_{t-a}$; $\phi(z) = 1 - \phi_1 z - \cdots - \phi_p z^p$, $\Phi(z) = 1 - \Phi_1 z - \cdots - \Phi_Q z^Q$; $\theta(z) = 1 - \theta_1 z - \cdots - \theta_q z^q$, $\Phi(z) = 1 - \Phi_1 z - \cdots - \Phi_Q z^Q$; $\varepsilon_t$ is identically and normally distributed with mean zero, variance $\sigma^2$; and $\text{cov}(\varepsilon_t, \varepsilon_{t-k}) = 0$, $\forall k \neq 0$, that is, $\{\varepsilon_t\} \sim WN(0,\sigma^2)$.

In the seasonal $ARIMA(p,d,q)(P,D,Q)_S$ model, the parameters $p$ and $P$ denote the nonseasonal and seasonal autoregressive polynomial order, respectively, and the parameters $q$ and $Q$ represent the nonseasonal and seasonal moving average polynomial order, respectively. As discussed above, the parameter $d$ is the order of normal differencing, and the parameter $D$ is the order of seasonal differencing. From a practical perspective, fitted seasonal ARIMA models provide linear state transition equations that can be applied recursively to produce single and multiple interval forecasts. Furthermore, seasonal ARIMA models can be readily expressed in state space form, thereby allowing adaptive Kalman filtering techniques to be employed to provide a self-tuning forecast model.

## 3.3 Multilayer perceptron network

The multilayer perceptron (MLP) network creates a complex mapping from inputs into appropriate outputs and thus enables the network to approximate almost any nonlinear function, even with one hidden layer. The relationship between the inputs ($y_{t-1}, y_{t-2}, \cdots, y_{t-p}$) and the output ($y_t$) has the following form:

$$y_t = \alpha_0 + \sum_{j=1}^{q} \alpha_j g\left(\beta_{oj} + \sum_{i=1}^{p} \beta_{ij} y_{t-i}\right) + \varepsilon_t \qquad (8)$$

where $\alpha_j$ and $\beta_{ij}$ are the network parameters and $p$ and $q$ are the number of input nodes and hidden nodes, respectively. The activation function of the hidden layer uses the logistic function $g(y) = 1/(1+\exp(-y))$ in this study.

Backpropagation (BP) algorithms are one of the most commonly used training algorithms for MLP networks that minimize the total square errors of in-sample forecasting results. One challenge is to determine the number of neurons in each layer, the number of hidden layers, momentum parameters and learning rates. To explore the optimal architecture of MLP networks, these parameters can be determined by means of the trial-and-error method or particle swarm optimization algorithms. Underlying economic theory can be used to help determine the optimal input size. In this study, we use the autoregressive model to identify the input size.

## 3.4 Long short-term memory network

The long short-term memory (LSTM) neural network proposed by Hochreiter and Schmidhuber (1997) is a special kind of recurrent neural network. The core components of the LSTM network are to use memory cells and gates to store information for long periods of time or to forget unnecessary information. LSTM neural networks have stable and powerful capabilities in solving long-term and short-term dependency issues. The key parameter of the LSTM neural network is the memory cell, which can memorize the temporal state. Hence, the LSTM neural network can add or remove information to the cell state by the input gate, forget gate and output gate. The basic calculation steps of the LSTM neural network can be expressed as follows:

1) The input gate controls the input activations. When new input information comes, if the input gate is activated, the new input information can be accumulated to the memory cell.
2) The forget gate can forget unnecessary information; if the forget gate is activated, the past memory cell status can be forgotten in the process.
3) If the output gate is activated, the latest memory cell output can be propagated to the ultimate state.

In this study, the LSTM neural network includes three layers: one input layer, one hidden layer and one output layer. We define $x = (x_1, x_2, \ldots, x_T)$ as the historical input data and $y = (y_1, y_2, \ldots, y_T)$ as the output data. Then, the predicted metro passenger flow

can be calculated by the following equations:

$$i_t = \sigma(W_{ix}x_t + W_{im}m_{t-1} + W_{ic}c_{t-1} + b_i) \tag{9}$$

$$f_t = \sigma(W_{fx}x_t + W_{fm}m_{t-1} + W_{fc}c_{t-1} + b_f) \tag{10}$$

$$c_t = f_t \circ c_{t-1} + i_t \circ g(W_{cx}x_t + W_{cm}m_{t-1} + b_c) \tag{11}$$

$$o_t = \sigma(W_{ox}x_t + W_{om}m_{t-1} + W_{oc}c_t + b_o) \tag{12}$$

$$m_t = o_t \circ h(c_t) \tag{13}$$

$$y_t = W_{ym}m_t + b_y \tag{14}$$

where $i_t$ represents the input gate, $f_t$ represents the forget gate, $c_t$ represents the activation vectors for each memory cell, $o_t$ represents the output gate, $m_t$ represents the activation vectors for each memory block, $W$ represents the weigh matrices, $b$ represents the bias vectors and $\circ$ represents the scalar product of two vectors.

$\sigma(\cdot)$ represents the standard logistics sigmoid function as follows:

$$\sigma(x) = \frac{1}{1+e^{-x}} \tag{15}$$

$g(\cdot)$ represents the centered logistic sigmoid function as follows:

$$g(x) = \frac{4}{1+e^{-x}} - 2 \quad x \in [-2,2] \tag{16}$$

$h(\cdot)$ represents the centered logistic sigmoid function as follows:

$$h(x) = \frac{2}{1+e^{-x}} - 1 \quad x \in [-1,1] \tag{17}$$

The hyperparameters of the LSTM network are trained based on the backpropagation algorithm. The objective function of the LSTM network is to minimize the mean squared error of the in-sample dataset. Due to the extensive mathematical derivations, the detailed execution steps are not covered in this section. Interesting readers may refer to Hochreiter and Schmidhuber (1997) for more information.

### 3.5 The framework of the AdaEnsemble learning approach

Traditional traffic flow forecasting methods assume constant variance of the data and forecast the current value as a function of its past values. An alternative way is to treat the traffic characteristics as a combination of cyclic, deterministic and volatile components that are determined by specific road conditions, regular traffic demand (commuters), traffic regulations (speed limit), and irregular components affected by traffic incidents, weather, and some other exogenous factors. Accurate and reliable traffic flow forecasting relies on a better understanding of the overall underlying components. Therefore, according to the discussion in the introduction, we assume that traffic flow is composed of three components: a periodic trend, a deterministic part, and a volatility part. The structure of the proposed model is the sum of the periodic trend, the deterministic part and the volatility:

$$x_t = p_t + d_t + v_t \tag{18}$$

where $x_t$ is the observed metro passenger flow during time $t$, $p_t$ is the periodic trend expressed as regression of the present on periodic sines and cosines, $d_t$ is the deterministic part of the metro passenger flow data after removing the periodic components, and $v_t$ is the volatility part of $x_t$.

The SARIMA model provides the required framework to highlight the cyclical patterns in the traffic flow data, and regression on the periodic trend reproduces the cyclic patterns. By removing these periodicities in the data, the residual parts of the traffic flows could be fitted by regression on its past long memory values. The LSTM neural network is introduced to fit the deterministic part of the passenger flow data. The volatility part of metro passenger flow could be modeled by regression on its past short-term memory values. The MLP network is employed to model the volatile part of the metro passenger flow data. **Fig. 1** provides a flowchart of our proposed AdaEnsemble learning approach.

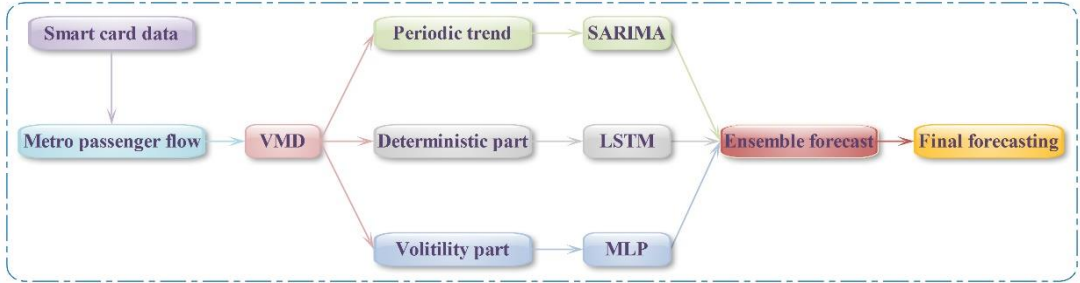

Fig. 1 The flowchart of the AdaEnsemble learning approach.

## 4 Empirical study

In this section, there are two main goals: (1) to evaluate the performance of our proposed AdaEnsemble learning approach for metro passenger flow forecasting and (2) to demonstrate the superiority of our proposed AdaEnsemble learning approach in comparison with several other benchmark models. To accomplish these two tasks, we collect smart card data from the Shenzhen metro system to test the forecasting performance of our proposed AdaEnsemble learning approach. The research data and evaluation criteria are introduced in **Section 4.1,** and the empirical results are analyzed in **Section 4.2**.

### 4.1 Data description and evaluation criteria

In this study, our proposed AdaEnsemble learning approach was applied to smart card data collected from the Shenzhen metro as a case study. The Shenzhen metro network expanded from 4 lines with 114 kilometers in 2006 to 16 lines with 442 kilometers in 2012 and led to a sudden increase in daily ridership from 1.93 million to 6.74 million. Among these metro stations, the Hui-Zhan-Zhong-Xin (HZZX) station, Fu-Ming (FM) station and Gang-Xia (GX) station are the three most representative stations with high passenger demands in the Shenzhen metro system. Hence, the metro passenger flows used in this study were collected from these three stations and aggregated into 15-min time intervals from transit smart cards for the HZZX, FM and GX subway stations between Oct. 14, 2013 and Nov. 30, 2013. For these stations, the service time of the subway stations is from 6:30 to 24:00. Because of the different passenger flow patterns between weekdays and weekends, the metro passenger flow data were divided into weekdays and weekends (Ke et al., 2017). The weekday and weekend data of the first two-thirds were selected as the in-sample dataset, and the remaining one-third of the data were selected as the out-of-sample dataset.

**Table 1** shows the descriptive statistics of the metro passenger flow data. This clearly indicates the difference in the statistical features among the datasets. For the three metro stations, the metro passenger flow data still have a sharp peak and a fatter tail. This characteristic indicates that the data do not satisfy the normal distribution but satisfy the leptokurtic $t$ distribution. The detailed data are not listed here but can be obtained from the authors.

**Table 1** Statistic characteristics of subway passenger flow.

| Stations | Type | Mean | Std.* | Skewness | Kurtosis |
|---|---|---|---|---|---|
| HZZX | Weekdays | 875.2556 | 686.8340 | 1.4902 | 4.9556 |
| | Weekends | 657.6381 | 369.6277 | 0.2045 | 2.7377 |
| FM | Weekdays | 549.5626 | 282.5867 | 0.8630 | 3.8464 |
| | Weekends | 506.4550 | 200.5142 | -0.6482 | 3.3894 |
| GX | Weekdays | 726.4776 | 517.3184 | 1.5130 | 4.9269 |
| | Weekends | 500.1788 | 223.3754 | 0.2427 | 4.5736 |

Notes: HZZX denotes Huizhanzhongxin station; FM denotes Fumin station; GX denotes Gangxia station. Std.* refers to the standard deviation.

Additionally, to compare the forecasting performance of our proposed AdaEnsemble learning approach with several other benchmark models, two evaluation criteria, namely, the mean absolute percentage error (MAPE) and root mean square error (RMSE), were employed to evaluate the forecasting performance of the in-sample dataset and out-of-sample dataset:

$$MAPE = \frac{1}{N}\sum_{i=1}^{N}\left|\frac{y_i - \hat{y}_i}{y_i}\right| \times 100\% \qquad (19)$$

$$RMSE = \left(\sum_{i=1}^{N}\frac{(y_i - \hat{y}_i)^2}{N}\right)^{1/2} \qquad (20)$$

where $\hat{y}_i$ and $y_i$ denote the forecasted and actual metro passenger flow at time $t$, and

$N$ is the number of observation samples. MAPE and RMSE measure the deviation between the actual and forecasted values, with smaller values indicating higher accuracy.

## 4.2 Empirical results

To verify the superiority of our proposed AdaEnsemble learning approach, five forecasting models are built and used as benchmarks (i.e., three single models, including the seasonal autoregressive integrated moving averaging (SARIMA) model, multilayer perceptron (MLP) neural network, and long short-term memory (LSTM) network), and two decomposition ensemble learning approaches, including VMD-MLP and VMD-LSTM. The reasons for choosing these benchmarks are as follows: (1) The SARIMA model has a noticeable impact on metro passenger flow forecasting as one of the periodical and seasonal models introduced in the econometrics literature and has shown its capacity in forecasting metro passenger flows (Smith et al., 2002). (2) The MLP and LSTM techniques are the most widely used neural networks in metro passenger flow forecasting, as introduced in **Section 1**. (3) The VMD-MLP and VMD-LSTM decomposition ensemble approaches verify the capability of adaptive modeling in our proposed approach.

The parameters of the SARIMA model are estimated by means of an automatic model selection algorithm implemented using the "forecast" program package in R software. For the MLP model, the number of inputs is determined using the partial mutual information method (maximum embedding order $d$=24). The number of outputs is set to one, and the number of hidden nodes (varying from 4 to 15) is determined by trial-and-error experiments. The logistic sigmoid function is selected as the activation function, and the backpropagation algorithm is employed to train the MLP. The MLP is implemented by the neural network toolbox in MATLAB 2017a software. Regarding the VMD algorithm, the optimal mode number is set to 3 using the difference between the center frequencies of the adjacent subseries, as the center frequency is closely related to the decomposition results of VMD (Dragomiretskiy and Zosso, 2014). The VMD algorithm is implemented using the VMD package in MATLAB 2017a software. For the LSTM neural network, the number of input nodes is determined using the partial mutual information method (maximum embedding order $d$=24). The number of output nodes is set to one, and the number of hidden layers is set to one. The number of hidden nodes (varying from 4 to 25) is determined by trial-and-error experiments. The LSTM is implemented using the LSTM package in the MATLAB 2017a computing environment.

Using the research design mentioned above, forecasting experiments were performed for metro passenger flow. Accordingly, the forecasting performance of all of the examined models is evaluated using the two accuracy measures.

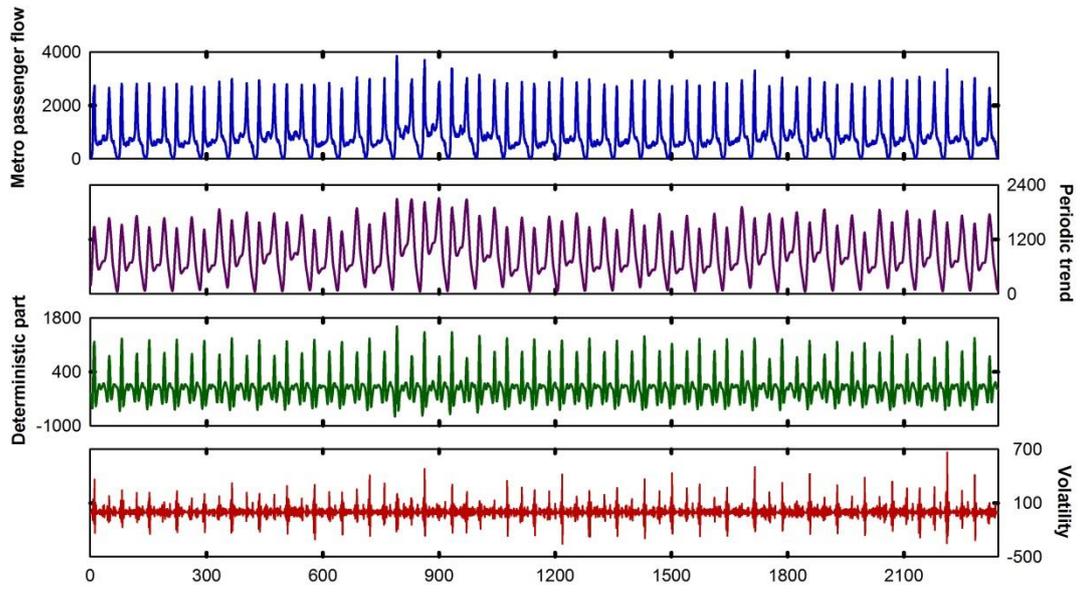

**Fig. 2** Decomposition of the weekday passenger flow data at the HZZX station.

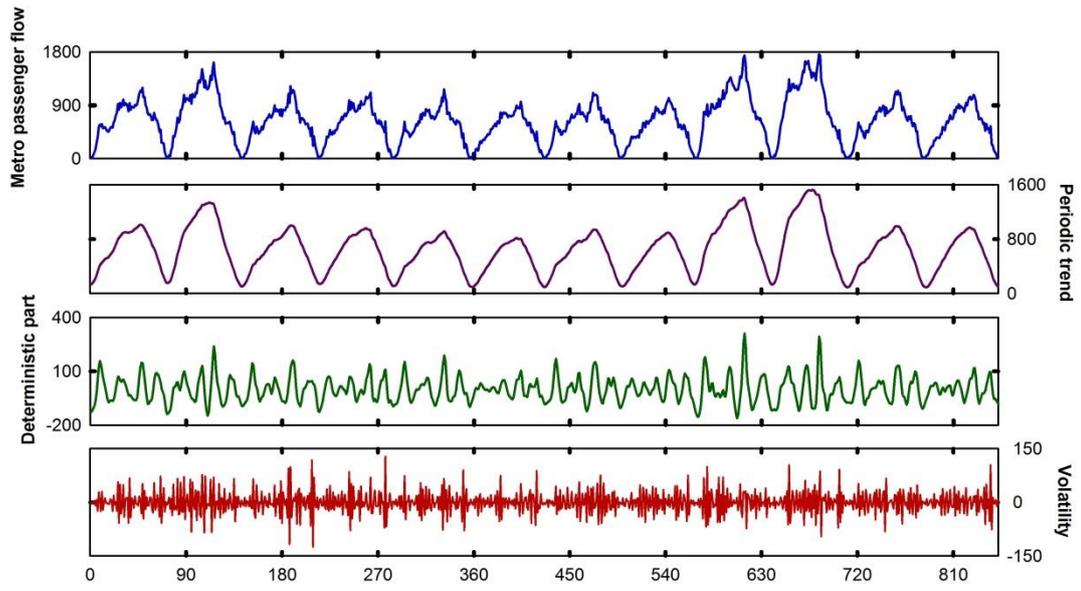

**Fig. 3** Decomposition of the weekend passenger flow data at the HZZX station.

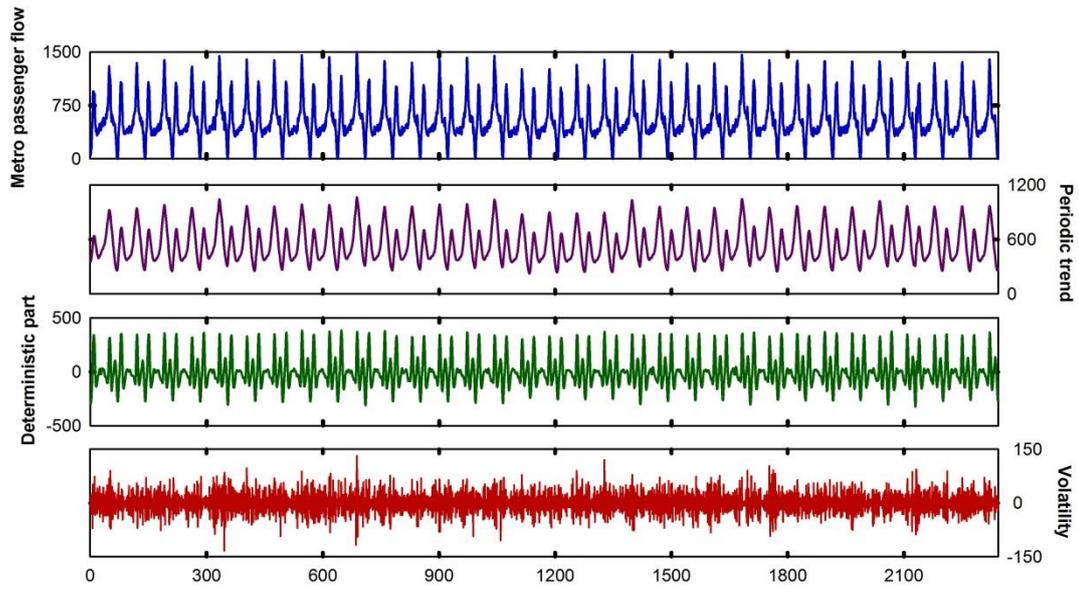

**Fig. 4** Decomposition of the weekday passenger flow data at the FM station.

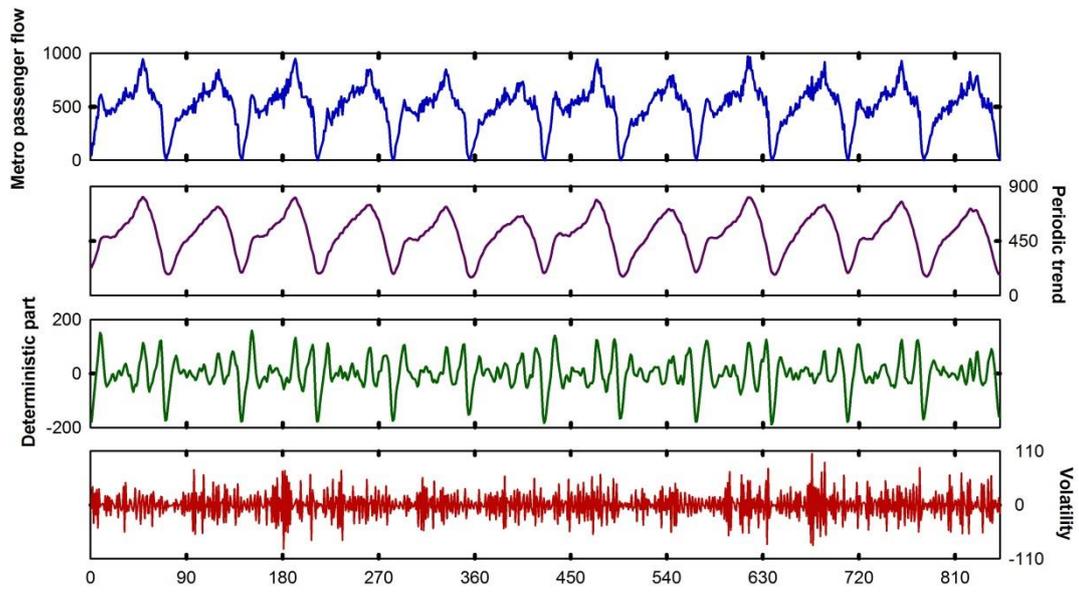

**Fig. 5** Decomposition of the weekend passenger flow data at the FM station.

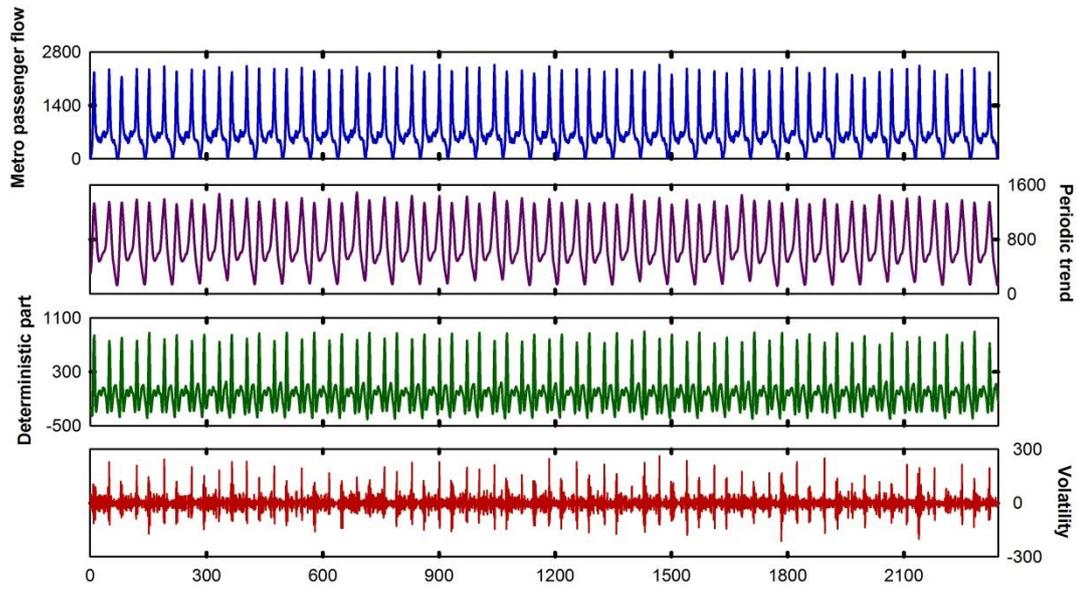

**Fig. 6** Decomposition of the weekday passenger flow data at the GX station.

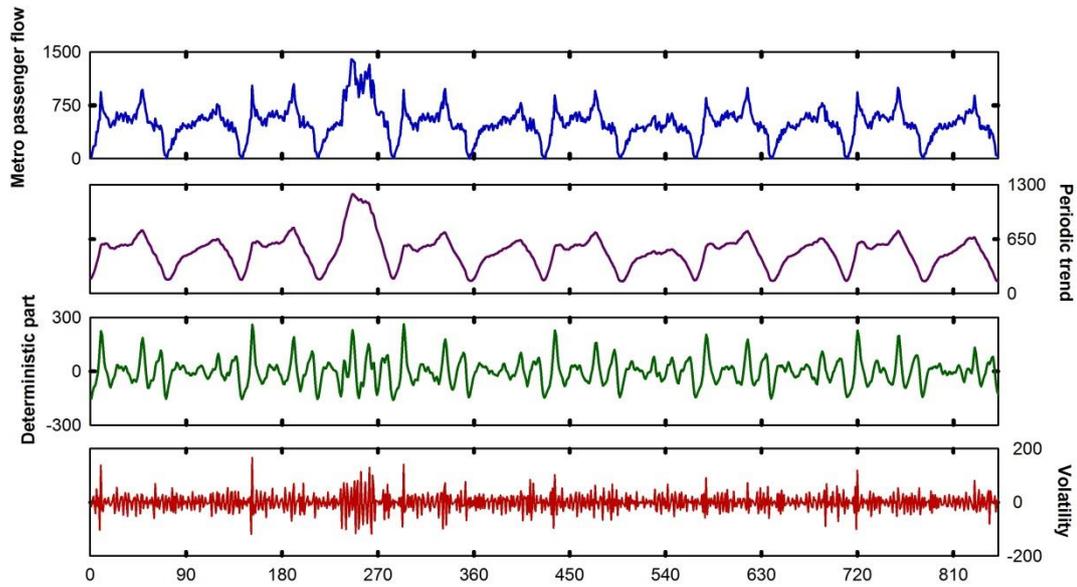

**Fig. 7** Decomposition of the weekend passenger flow data at the GX station.

**Table 2** Measures of each component for weekdays and weekends in metro passenger flows at three stations.

| Stations | Modes | Weekdays | | | Weekends | | |
|---|---|---|---|---|---|---|---|
| | | Mean period | Correlation coefficient | Variance as % of observed | Mean period | Correlation coefficient | Variance as % of observed |
| HZZX | Volatility | 2.96 | 0.19 | 1.12 | 2.22 | 0.12 | 0.76 |
| | Deterministic part | 35.50 | 0.76 | 26.46 | 17.75 | 0.43 | 3.67 |
| | Periodic trend | 71.00 | 0.86 | 46.20 | 71.00 | 0.97 | 86.26 |
| FM | Volatility | 2.45 | 0.20 | 1.26 | 2.15 | 0.17 | 1.80 |
| | Deterministic part | 28.40 | 0.78 | 23.31 | 17.75 | 0.57 | 8.91 |
| | Periodic trend | 71.00 | 0.89 | 46.01 | 71.00 | 0.95 | 72.53 |
| | Volatility | 2.73 | 0.17 | 0.95 | 3.94 | 0.21 | 2.20 |
| GX | Deterministic part | 28.40 | 0.78 | 26.11 | 17.75 | 0.58 | 9.95 |
| | Periodic trend | 71.00 | 0.87 | 44.45 | 71.00 | 0.94 | 70.72 |

The decomposition results of the weekday and weekend passenger flow series at the three metro stations using VMD are shown in **Figs. 2-7**. We note that each original passenger flow dataset is decomposed into periodic, deterministic and volatile components through the VMD algorithm. All of the periodic components of these metro passenger flow series show a one-day cycle. Additionally, the following measures are considered when analyzing each component, such as the mean period of each component, the correlation coefficient between the original passenger flow series and each component, and the variance percentage of each component. **Table 2** presents the measures of each component for the weekday and weekend metro passenger flows at the three stations. The mean period under study is defined as the value obtained by dividing the total number of points by the peak number of each component, because the amplitude and frequency of a component may change continuously with time and the period is not constant. The Pearson correlation coefficient is used to measure the correlations between the original passenger flow series and each component. However, because these components are independent of each other, it may be possible to use the variance percentage to explain the contribution of each component to the total volatility of the observed passenger flow series. The results of all six decompositions show that the dominant mode of the observed data is not volatility and deterministic parts but the periodic trend. For all the weekday metro passenger flow decompositions, the coefficients between the original passenger flow series and periodic component reach 0.86, 0.89 and 0.87 for the HZZX, FM and GX stations, respectively. However, for all the weekend metro passenger flow decompositions, the coefficients between the original passenger flow series and periodic component reach high levels of more than 0.97, 0.95 and 0.94 for the HZZX, FM and GX stations, respectively. Moreover, the variance of the periodic component accounts for more than 45% of the total volatility of the observed passenger flow data. The highest value is more than 86%.

After the decomposition, as discussed in **Section 3.5**, the SARIMA model is used to forecast the extracted periodic component, the LSTM neural network is employed to forecast the extracted deterministic component, and the MLP neural network is used to forecast the extracted volatile component. Finally, the forecasting results of the periodic, deterministic and volatile components are integrated into an aggregated output via another MLP neural network.

The forecasting performance of the six models (i.e., AdaEnsemble, VMD-LSTM, VMD-MLP, LSTM, MLP, and SARIMA) under study at the three stations across the ten forecasting horizons ($h$-step-ahead, i.e., $h$=1, 2, …, 10) for RMSE and MAPE are shown in **Tables 3-8**.

**Table 3**
The RMSE values of different forecasting models at the HZZX station during weekdays and weekends.

| Types | RMSE | Number of forecasting steps ahead |
|---|---|---|

|  |  | 1 | 2 | 3 | 4 | 5 | 6 | 7 | 8 | 9 | 10 |
|---|---|---|---|---|---|---|---|---|---|---|---|
| Weekdays | SARIMA | 152.91 | 158.26 | 160.07 | 162.25 | 164.73 | 166.81 | 169.78 | 173.44 | 175.08 | 178.31 |
|  | MLP | 132.47 | 133.45 | 135.68 | 137.22 | 139.58 | 142.06 | 146.03 | 150.26 | 154.23 | 158.27 |
|  | LSTM | 104.35 | 107.07 | 109.13 | 110.94 | 112.37 | 115.63 | 119.68 | 121.39 | 125.69 | 129.45 |
|  | VMD-MLP | 78.88 | 80.61 | 82.62 | 84.67 | 86.61 | 88.59 | 91.67 | 93.45 | 96.75 | 97.26 |
|  | VMD-LSTM | 75.69 | 79.06 | 81.06 | 83.74 | 85.69 | 86.79 | 88.13 | 90.22 | 93.06 | 94.69 |
|  | AdaEnsemble | 60.25 | 61.34 | 63.61 | 65.18 | 67.08 | 68.34 | 70.43 | 73.16 | 75.22 | 76.35 |
| Weekends | SARIMA | 142.68 | 143.58 | 144.67 | 146.35 | 147.11 | 146.24 | 148.37 | 152.08 | 155.34 | 159.74 |
|  | MLP | 104.16 | 105.39 | 105.68 | 107.94 | 112.69 | 115.71 | 118.24 | 121.35 | 125.68 | 130.16 |
|  | LSTM | 89.84 | 91.25 | 93.51 | 93.18 | 94.69 | 97.17 | 101.27 | 105.38 | 109.32 | 113.53 |
|  | VMD-MLP | 72.38 | 74.06 | 75.67 | 77.38 | 79.25 | 82.34 | 85.61 | 86.92 | 87.16 | 90.75 |
|  | VMD-LSTM | 68.59 | 69.17 | 71.98 | 73.46 | 76.15 | 78.42 | 80.19 | 83.67 | 84.91 | 86.24 |
|  | AdaEnsemble | 49.72 | 50.24 | 52.56 | 53.69 | 55.18 | 57.88 | 60.17 | 63.35 | 65.04 | 68.43 |

**Table 4**

The MAPE values of different forecasting models at the HZZX station during weekdays and weekends.

| Types | MAPE (%) | Number of forecasting steps ahead | | | | | | | | | |
|---|---|---|---|---|---|---|---|---|---|---|---|
|  |  | 1 | 2 | 3 | 4 | 5 | 6 | 7 | 8 | 9 | 10 |
| Weekdays | SARIMA | 16.84 | 16.98 | 17.58 | 17.86 | 18.01 | 18.54 | 18.43 | 19.35 | 19.58 | 20.67 |
|  | MLP | 10.11 | 10.56 | 11.36 | 12.87 | 13.79 | 14.88 | 13.93 | 15.61 | 16.05 | 16.36 |
|  | LSTM | 8.52 | 8.87 | 9.53 | 9.38 | 10.58 | 10.69 | 11.37 | 12.54 | 12.39 | 13.24 |
|  | VMD-MLP | 4.91 | 5.29 | 6.35 | 6.86 | 7.56 | 8.39 | 9.43 | 8.88 | 10.38 | 11.37 |
|  | VMD-LSTM | 4.76 | 5.03 | 5.56 | 6.43 | 7.06 | 6.88 | 7.46 | 8.25 | 9.43 | 10.69 |
|  | AdaEnsemble | 3.14 | 3.35 | 3.69 | 4.12 | 4.36 | 4.18 | 5.68 | 6.81 | 7.33 | 8.12 |
| Weekends | SARIMA | 14.69 | 15.02 | 15.62 | 16.08 | 16.54 | 17.82 | 18.33 | 19.46 | 20.16 | 21.46 |
|  | MLP | 8.74 | 8.66 | 9.43 | 9.58 | 10.39 | 11.64 | 12.38 | 13.45 | 13.82 | 14.57 |
|  | LSTM | 6.91 | 7.16 | 7.58 | 8.06 | 8.97 | 9.58 | 9.16 | 10.35 | 10.62 | 11.19 |
|  | VMD-MLP | 4.62 | 4.55 | 5.43 | 5.87 | 6.25 | 6.87 | 7.16 | 7.58 | 8.59 | 9.28 |
|  | VMD-LSTM | 4.05 | 4.35 | 5.02 | 5.36 | 5.88 | 6.13 | 6.59 | 7.06 | 8.12 | 8.91 |
|  | AdaEnsemble | 2.74 | 3.16 | 3.87 | 3.69 | 4.51 | 4.55 | 5.02 | 5.46 | 5.97 | 6.32 |

**Table 5**

The RMSE values of different forecasting models at the FM station during weekdays and weekends.

| Types | RMSE | Number of forecasting steps ahead | | | | | | | | | |
|---|---|---|---|---|---|---|---|---|---|---|---|
|  |  | 1 | 2 | 3 | 4 | 5 | 6 | 7 | 8 | 9 | 10 |
| Weekdays | SARIMA | 105.81 | 105.87 | 106.54 | 107.33 | 107.89 | 109.25 | 110.78 | 110.74 | 111.69 | 112.16 |
|  | MLP | 76.83 | 77.89 | 79.57 | 81.33 | 82.46 | 85.57 | 87.39 | 90.88 | 89.42 | 95.51 |
|  | LSTM | 63.81 | 65.29 | 65.74 | 66.16 | 67.89 | 68.57 | 70.25 | 72.33 | 75.87 | 79.06 |
|  | VMD-MLP | 48.98 | 49.35 | 48.87 | 50.56 | 51.28 | 52.37 | 51.89 | 53.45 | 54.88 | 55.71 |
|  | VMD-LSTM | 39.37 | 40.25 | 44.36 | 50.25 | 49.88 | 50.69 | 51.83 | 52.15 | 51.76 | 52.46 |
|  | AdaEnsemble | 27.85 | 28.31 | 27.56 | 29.09 | 29.54 | 30.25 | 31.06 | 31.47 | 32.87 | 33.23 |
| Weekends | SARIMA | 100.47 | 100.37 | 101.69 | 101.28 | 101.96 | 103.57 | 104.68 | 105.37 | 107.49 | 109.62 |
|  | MLP | 79.46 | 80.43 | 80.65 | 81.55 | 81.57 | 81.49 | 82.61 | 82.54 | 83.43 | 84.73 |
|  | LSTM | 59.58 | 60.39 | 61.33 | 60.17 | 61.22 | 62.41 | 63.87 | 63.68 | 64.06 | 65.39 |

| | | | | | | | | | | |
|---|---|---|---|---|---|---|---|---|---|---|
| | VMD-MLP | 44.87 | 45.27 | 46.31 | 45.56 | 46.29 | 47.34 | 48.58 | 50.11 | 49.26 | 50.27 |
| | VMD-LSTM | 37.69 | 37.54 | 38.26 | 39.13 | 39.25 | 40.41 | 40.46 | 41.33 | 42.29 | 43.14 |
| | AdaEnsemble | 26.48 | 26.69 | 27.33 | 28.45 | 28.63 | 27.56 | 28.53 | 29.67 | 30.31 | 30.88 |

**Table 6**

The MAPE values of different forecasting models at the FM station during weekdays and weekends.

| Types | MAPE | Number of forecasting steps ahead | | | | | | | | | |
|---|---|---|---|---|---|---|---|---|---|---|---|
| | | 1 | 2 | 3 | 4 | 5 | 6 | 7 | 8 | 9 | 10 |
| Weekdays | SARIMA | 18.04 | 19.15 | 19.27 | 19.51 | 19.62 | 19.83 | 20.14 | 20.13 | 21.13 | 21.58 |
| | MLP | 12.94 | 13.54 | 13.87 | 14.02 | 14.53 | 15.07 | 15.51 | 16.14 | 15.73 | 16.37 |
| | LSTM | 10.46 | 11.07 | 11.35 | 12.03 | 12.32 | 12.45 | 12.67 | 13.16 | 13.74 | 14.26 |
| | VMD-MLP | 8.02 | 8.07 | 8.05 | 8.31 | 8.33 | 8.74 | 8.56 | 9.34 | 9.49 | 10.11 |
| | VMD-LSTM | 6.13 | 6.32 | 7.07 | 8.14 | 8.15 | 8.26 | 8.45 | 8.68 | 9.01 | 9.12 |
| | AdaEnsemble | 4.07 | 4.35 | 4.67 | 4.79 | 4.92 | 5.32 | 5.61 | 5.66 | 5.65 | 6.05 |
| Weekends | SARIMA | 15.27 | 16.01 | 16.59 | 16.07 | 17.05 | 17.34 | 17.56 | 17.68 | 18.29 | 18.57 |
| | MLP | 12.56 | 13.03 | 13.28 | 13.89 | 13.15 | 13.67 | 14.56 | 14.69 | 15.13 | 15.47 |
| | LSTM | 9.41 | 9.68 | 10.16 | 9.97 | 10.23 | 10.86 | 11.02 | 11.14 | 11.67 | 11.93 |
| | VMD-MLP | 7.59 | 7.67 | 7.72 | 7.60 | 7.78 | 8.06 | 8.10 | 8.35 | 8.46 | 8.68 |
| | VMD-LSTM | 6.39 | 6.41 | 6.38 | 6.54 | 6.66 | 6.76 | 6.89 | 7.25 | 7.44 | 7.78 |
| | AdaEnsemble | 4.41 | 4.46 | 4.55 | 4.76 | 4.77 | 4.81 | 4.83 | 5.06 | 5.21 | 5.22 |

**Table 7**

The RMSE values of different forecasting models in the GX station during weekdays and weekends.

| Types | RMSE | Number of forecasting steps ahead | | | | | | | | | |
|---|---|---|---|---|---|---|---|---|---|---|---|
| | | 1 | 2 | 3 | 4 | 5 | 6 | 7 | 8 | 9 | 10 |
| Weekdays | SARIMA | 91.26 | 92.36 | 93.58 | 95.02 | 95.96 | 97.43 | 98.57 | 99.28 | 101.26 | 102.33 |
| | MLP | 71.43 | 72.38 | 73.44 | 74.16 | 73.98 | 75.13 | 75.87 | 76.59 | 78.18 | 80.16 |
| | LSTM | 59.16 | 60.89 | 62.13 | 61.44 | 63.57 | 65.06 | 67.13 | 67.86 | 68.59 | 69.74 |
| | VMD-MLP | 45.68 | 46.58 | 48.26 | 49.13 | 50.17 | 51.25 | 53.06 | 52.88 | 54.69 | 56.07 |
| | VMD-LSTM | 39.41 | 40.33 | 41.58 | 43.54 | 44.06 | 44.96 | 45.16 | 46.68 | 48.27 | 49.31 |
| | AdaEnsemble | 20.25 | 20.84 | 21.26 | 22.53 | 22.07 | 22.95 | 23.18 | 23.49 | 25.41 | 26.05 |
| Weekends | SARIMA | 85.78 | 86.47 | 87.25 | 87.92 | 86.36 | 88.64 | 89.13 | 90.06 | 91.25 | 93.36 |
| | MLP | 53.61 | 54.21 | 54.92 | 55.67 | 54.88 | 56.07 | 57.19 | 58.47 | 60.06 | 62.37 |
| | LSTM | 45.56 | 46.61 | 47.29 | 46.58 | 50.36 | 51.47 | 52.38 | 54.92 | 57.44 | 59.68 |
| | VMD-MLP | 37.89 | 37.71 | 39.43 | 40.11 | 40.88 | 41.29 | 42.06 | 43.18 | 44.68 | 45.16 |
| | VMD-LSTM | 34.25 | 34.68 | 35.62 | 36.09 | 36.85 | 37.53 | 38.41 | 38.67 | 39.16 | 40.02 |
| | AdaEnsemble | 20.21 | 21.07 | 22.31 | 21.97 | 23.45 | 24.62 | 23.58 | 25.16 | 25.43 | 26.51 |

**Table 8**

The MAPE values of different forecasting models at the GX station during weekdays and weekends.

| Types | MAPE | Number of forecasting steps ahead | | | | | | | | | |
|---|---|---|---|---|---|---|---|---|---|---|---|
| | | 1 | 2 | 3 | 4 | 5 | 6 | 7 | 8 | 9 | 10 |
| Weekdays | SARIMA | 16.30 | 16.49 | 16.71 | 16.97 | 17.13 | 17.40 | 17.61 | 17.73 | 18.08 | 18.27 |
| | MLP | 12.74 | 12.93 | 13.11 | 13.24 | 13.21 | 13.42 | 13.54 | 13.68 | 13.96 | 14.31 |
| | LSTM | 10.35 | 10.88 | 11.09 | 10.97 | 11.36 | 11.68 | 12.01 | 12.12 | 12.25 | 12.45 |
| | VMD-MLP | 7.89 | 8.31 | 8.62 | 8.77 | 9.02 | 9.16 | 9.53 | 9.49 | 9.77 | 10.01 |
|          | VMD-LSTM    | 6.91  | 6.95  | 7.17  | 7.51  | 7.60  | 7.75  | 7.79  | 8.05  | 8.96  | 9.05  |
|          | AdaEnsemble | 3.55  | 3.59  | 3.67  | 3.85  | 4.02  | 4.19  | 5.04  | 5.68  | 5.39  | 6.01  |
|          | SARIMA      | 15.40 | 15.67 | 15.89 | 16.01 | 16.29 | 15.55 | 16.78 | 17.02 | 16.94 | 17.56 |
|          | MLP         | 9.41  | 9.51  | 9.63  | 9.77  | 9.62  | 9.84  | 10.06 | 10.59 | 10.53 | 11.04 |
| Weekends | LSTM        | 8.18  | 8.18  | 8.31  | 8.17  | 8.84  | 9.03  | 9.19  | 9.64  | 10.08 | 10.47 |
|          | VMD-MLP     | 6.64  | 6.62  | 6.92  | 7.04  | 7.17  | 7.26  | 7.38  | 7.58  | 7.84  | 8.16  |
|          | VMD-LSTM    | 6.01  | 6.08  | 6.25  | 6.33  | 6.47  | 6.59  | 6.74  | 6.78  | 6.87  | 7.02  |
|          | AdaEnsemble | 3.63  | 3.69  | 3.92  | 3.85  | 4.12  | 4.33  | 4.14  | 4.41  | 4.46  | 4.65  |

The results in the above tables show that our proposed AdaEnsemble approach is the best one for metro passenger flow forecasting among all forecasting horizons ($h$-step-ahead, i.e., $h$=1, 2, …, 10) for the three metro stations compared with the other five benchmarks under study. It is conceivable that the reason behind the inferiority of the LSTM and MLP relative to the AdaEnsemble approach is that the two pure neural networks cannot model periodic components directly. Therefore, prior data processing, such as time series decomposition, is critical and necessary to build a better forecaster, which is implemented as our proposed AdaEnsemble approach under study.

Additionally, from the results of all models under study, the SARIMA model is consistently the worst forecaster for each metro passenger flow in terms of forecasting accuracy and horizons. It is conceivable that the reason behind the inferiority of the SARIMA is that it is a typical linear model and cannot capture nonlinear patterns in metro passenger flows.

From the above analysis of the empirical results, several interesting findings can be drawn. (1) LSTM performs better than the single benchmark models. (2) In a comparison between VMD-LSTM (VMD-MLP) and LSTM (MLP), VMD-LSTM (VMD-MLP) is the winner. This means that mode decomposition of the metro passenger flow time series before further forecasting can effectively enhance the forecasting power for metro passenger flow forecasting. (3) Due to the highly nonlinear and periodic patterns in the metro passenger flow series, AI-based nonlinear models are more suitable for forecasting time series with highly periodic volatility than linear models. (4) Our proposed AdaEnsemble approach is consistently the best compared with other benchmarks under study for metro passenger flow forecasting by means of statistical accuracy and forecasting horizons. (5) Our proposed AdaEnsemble approach can be considered a promising solution for forecasting time series with highly periodic volatility.

## 5 Conclusions

In this research, we present a novel adaptive ensemble (AdaEnsemble) learning approach to accurately forecast the volume of metro passenger flows. This approach decomposes the time series of metro passenger flows into periodic components, deterministic components and volatility components by variational mode decomposition (VMD). Then, we employ the SARIMA model to forecast the periodic component, the LSTM network to learn and forecast the deterministic component and the MLP network to forecast the volatility component. In the last stage, the diverse forecasted components are reconstructed by another MLP network.

Due to the highly nonlinear and periodic patterns in the metro passenger flow series, the advantage of the proposed approach is that it decomposes the original data into periodic components, deterministic components, and volatility components and then employs suitable methods to predict the characteristics of diverse components. Finally, the diverse forecasted components are reconstructed by an MLP network. The empirical results show that (1) mode decomposition of the metro passenger flow time series before further forecasting can effectively enhance the forecasting power for metro passenger flow forecasting; (2) the hybrid model with linear models and nonlinear models is more suitable for forecasting time series with highly periodic volatility; and (3) our proposed AdaEnsemble learning approach has the best forecasting performance compared with the state-of-the-art models in terms of statistical accuracy and forecasting horizons.

The metro passenger flows are influenced by many factors, such as special events, extreme weather conditions, and accidents. Our proposed AdaEnsemble learning approach is a univariate and hybrid model, and it is difficult to accurately capture the uncertainty in the metro passenger flow. In a future study, we will try to address these issues and improve prediction accuracy by employing new methods, new variables or an integrated forecasting framework.

## Conflict of interests

The authors declare that there are no conflicts of interest regarding the publication of this paper.

## Acknowledgments

This research work was partly supported by the National Natural Science Foundation of China under Grants No. 71988101 and No. 71642006.